\documentclass[10pt,twocolumn,letterpaper]{article}

\usepackage[pagenumbers]{wacv} 

\definecolor{wacvblue}{rgb}{0.21,0.49,0.74}
\usepackage[pagebackref,breaklinks,colorlinks,allcolors=wacvblue]{hyperref}

\def\wacvPaperID{3117} 
\def\confName{WACV}
\def\confYear{2026}

\newcommand{\name}{KeyBench \xspace}

\usepackage{multirow}
\usepackage{algorithm}
\usepackage{algpseudocode}

\title{From Captions to Keyframes: KeyScore for Multimodal Frame Scoring and Video-Language Understanding}

\author{Shih-Yao Lin, Sibendu Paul, Caren Chen\\
Amazon Prime Video\\
{\tt\small {mikeslin, sibendu, carechen}@amazon.com}
}

\begin{document}
\maketitle

\begin{abstract}
Selecting informative keyframes is critical for efficient video understanding, yet existing approaches often rely on heuristics, ignore semantics, or produce redundant frames.  
We propose \textbf{KeyScore}, a caption-aware frame scoring method that combines three complementary signals: semantic similarity to captions, temporal representativeness, and contextual drop impact. 
Applied to large-scale video–caption datasets, KeyScore generates frame-level importance scores that enable training keyframe extractors or guiding video–language models. 
To support this, we also propose \textbf{STACFP}—a Spatio-Temporal Adaptive Clustering method that generates diverse and compact frame proposals across long videos.
Together, KeyScore and STACFP reduce uninformative frames while preserving critical content, resulting in faster and more accurate inference. 
Our experiments on three standard video-language benchmarks (MSRVTT, MSVD, DiDeMo) show that combining STACFP and KeyScore enables up to \textbf{99\% frame reduction} compared to full-frame processing, while \textbf{outperforming uniform 8-frame encoders} in video-text retrieval, keyframe extraction, and action recognition tasks. 
By focusing on semantically relevant frames, our method enhances both efficiency and performance, enabling scalable and caption-grounded video understanding.



\end{abstract}

\begin{figure}[t]
    \centering
    \includegraphics[width=\linewidth]{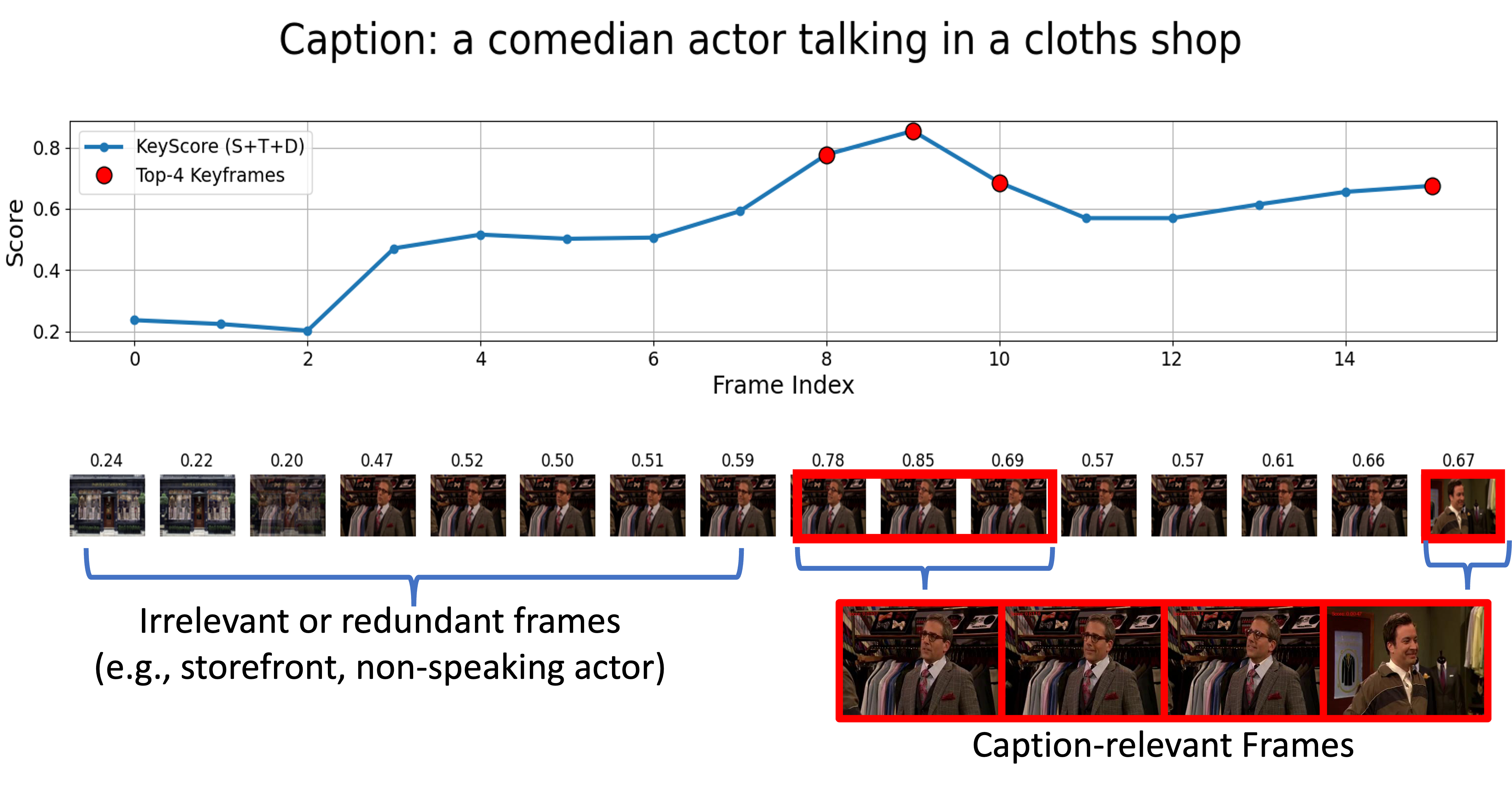}
    \vspace{-0.2in}
    \caption{
\textbf{Motivating example of our frame scoring.}  
Given the caption \textit{``a comedian actor talking in a cloths shop''}, our method selects keyframes that are semantically aligned with the caption (e.g., actor speaking), while avoiding irrelevant or repetitive frames (e.g., storefront, similar poses).
}
\label{fig:motivation}
\end{figure}

\section{Introduction}
\label{sec:intro}
With the exponential growth of video content, video understanding has become a central challenge in multimedia research, powering tasks such as video captioning~\cite{survey_video_captioning}, video-text retrieval~\cite{wu2023cap4video}, and action recognition~\cite{survey_action_recognition}. 
A persistent bottleneck across these domains is the need to process long, redundant, and often noisy frame sequences. Such inefficiency not only strains computation but also dilutes semantic signals. Selecting a compact yet informative set of keyframes—those that best capture the core content of a video—offers a promising path toward both efficiency and accuracy. Figure~\ref{fig:motivation} illustrates the goal of our caption-aware frame scoring approach: to highlight semantically relevant and diverse frames while suppressing those that are visually redundant or off-topic with respect to the caption.


Despite its importance, \textbf{keyframe scoring has been underexplored from a semantics-aware perspective}. Prior approaches~\cite{tang2023deep,dong2024m2,keyvideollm, Koala} often rely on low-level features, heuristics, or unsupervised clustering, which fail to capture high-level semantics in captions. A common fallback in video encoders and Video-LLMs is \textbf{uniform frame sampling}, which misses critical events, repeats redundant frames, and ignores semantics and motion. Clustering-based methods such as \textbf{SCFP} (e.g., Katna~\cite{katna}) improve diversity but neglect temporal dynamics, semantic grounding, and scene boundaries, while the choice of cluster number $k$ remains non-trivial and dataset-dependent.

To address these limitations at the proposal stage, we introduce \textbf{Spatio-Temporal Adaptive Clustering for Frame Proposals (STACFP)}, which augments clustering with temporal encoding and automatically selects the optimal number of clusters via silhouette analysis. Unlike SCFP, STACFP adaptively allocates more proposals to dynamic regions while avoiding redundancy in static segments, producing a compact yet diverse set of candidate frames that better reflect the temporal structure of the video.

On top of these proposals, we introduce \textbf{KeyScore}, a caption-aware frame scoring method designed to identify the most informative frames in video–language tasks. KeyScore integrates three complementary signals: 
(1) \textit{semantic similarity} between frames and captions, 
(2) \textit{temporal representativeness} to ensure coverage of the video timeline, and 
(3) \textit{contextual drop impact} to account for redundancy and diversity. 
Together, these signals provide frame-level importance scores that can guide keyframe extraction, improve the efficiency of video encoders, and accelerate inference in Video-LLMs.  

KeyScore offers two key advantages. First, it provides a flexible framework that can be applied directly to large-scale video–caption datasets, generating frame-level importance scores without requiring manual annotations. Second, it enables new evaluation paradigms where frame quality is judged by \textbf{semantic alignment and downstream task performance} rather than heuristics alone.  

We extensively validate KeyScore across retrieval (MSR-VTT, MSVD, DiDeMo), keyframe extraction (TVSum20, SumMe), and zero-shot action classification (HMDB-51). Results show that KeyScore consistently outperforms uniform sampling and clustering-based baselines, improving accuracy while reducing frame usage by up to 97–99\% compared to raw videos and 63–75\% compared to standard 8-frame encoders. These findings demonstrate that caption-aware frame scoring is a powerful tool for content-efficient video understanding.


\vspace{1mm}

\noindent \textbf{Our contributions are three-fold:}
\begin{itemize}
    \item We propose \textbf{KeyScore}, a caption-aware frame scoring method that integrates semantic relevance, temporal diversity, and drop impact to select keyframes aligned with video captions.
    
    \item We introduce \textbf{STACFP} (Spatio-Temporal Adaptive Clustering for Frame Proposals), a lightweight yet effective sampling strategy that selects diverse candidate frames while preserving important content.

    \item We show that KeyScore improves task performance while significantly reducing computational cost—achieving up to 99\% frame reduction compared to processing all frames, and outperforming standard sparse sampling strategies (e.g., uniform 8-frame inputs) by focusing on caption-relevant content and filtering out uninformative frames.

\end{itemize}

\vspace{1mm}

    



\section{Related Works}
\label{sec:related_works}

\subsection{Keyframe Selection and Video Summarization}
Keyframe selection and video summarization aim to extract the most informative or representative frames from a video, thereby reducing redundancy while preserving essential content. Traditional approaches rely on low-level features such as motion, color histograms, or temporal differences to identify representative or diverse frames~\cite{zhang2016summary, gong2014diverse, zhang2025tokenbinder}.
Katna~\cite{katna}, for instance, applies K-means clustering on frame histograms and selects the sharpest frame (via Laplacian variance) from each cluster, further filtering based on LUV color differences, brightness, and contrast. While effective, such methods are highly sensitive to feature design and hyperparameter tuning.

Recent learning-based methods have shifted toward supervised or unsupervised frame importance prediction using deep visual features~\cite{keyvideollm, tang2023deep, Koala, tang2025adaptive}. However, these approaches often lack semantic grounding from natural language annotations (e.g., captions), which limits their ability to select frames relevant to higher-level video-language tasks. Attention-based video transformers~\cite{timesformer} and reinforcement learning strategies~\cite{liu2022video} have also been explored, but a consistent limitation is the absence of standardized, semantically informed evaluation criteria—making comparisons across methods less meaningful.

\subsection{Frame Sampling and Proposal Methods}
Uniform sampling is widely used in Video-LLMs~\cite{vtimellm, videollava, videollama, qwen, qwen2} for its simplicity, but often overlooks dynamic moments and yields redundant frames in static regions.  

Clustering-based methods such as VSUMM~\cite{vsumm} and Katna~\cite{katna} improve diversity but ignore temporal structure and require predefining the number of clusters. Adaptive variants incorporate silhouette scores~\cite{ejaz2012adaptive} or use segmentation-based strategies such as KTS~\cite{afham2023revisiting}. LMSKE~\cite{LMSKE} applies per-shot clustering with vision-language features, while TSDPC~\cite{tang2023deep} leverages density peak clustering over temporal segments. Despite their improvements, these methods remain limited by their lack of semantic integration.  


In contrast, our \textbf{STACFP} sampler performs lightweight global spatio-temporal clustering with automatic $k$ selection, relying on scene transitions rather than caption information. This generates proposals that are temporally diverse and structurally coherent, establishing a strong foundation for subsequent caption-aware scoring and video-language tasks.

\begin{figure*}[t]
    \centering
    \includegraphics[width=0.9\linewidth]{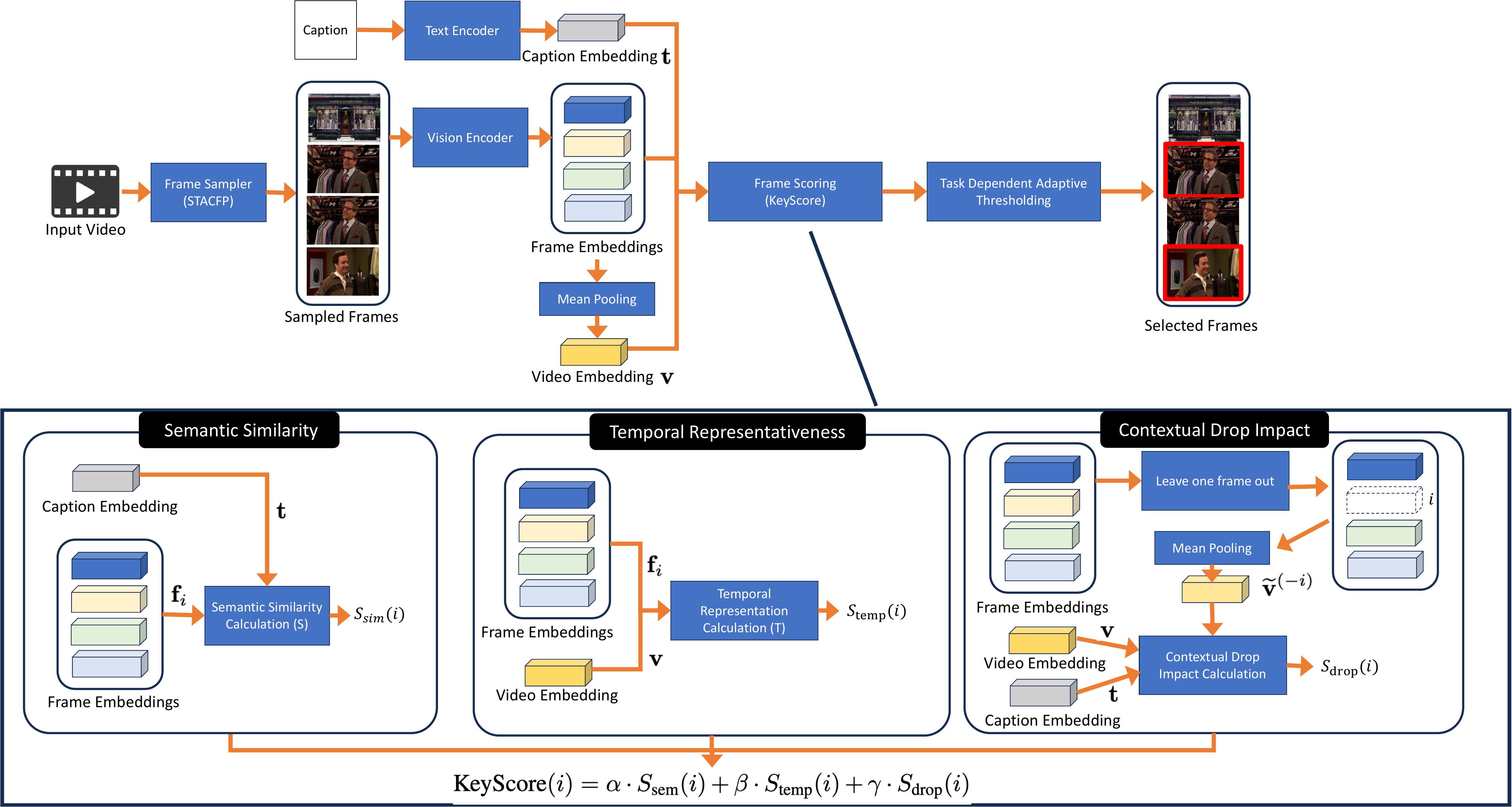}
    \caption{
    End-to-end pipeline of our proposed approach. STACFP first generates candidate keyframes from the input video. Caption and frame embeddings are then extracted using a text encoder and a vision encoder. The frame scoring module (\textbf{KeyScore}) integrates semantic similarity, temporal representation, and contextual drop impact to assign scores to each frame. Finally, task-dependent adaptive thresholding selects the most representative frames for downstream tasks such as retrieval, classification, or summarization.
    }
    \label{fig:pipeline}
\end{figure*}

\subsection{Semantic \& Embedding-Aware Frame Scoring}
With the rise of vision-language pretraining, frame selection has increasingly leveraged semantic alignment with text. KeyVideoLLM~\cite{keyvideollm} uses CLIP-based text–frame similarity to achieve high compression while enhancing video QA. AKS~\cite{tang2025adaptive} formulates keyframe selection as prompt-aware optimization, balancing semantic relevance with temporal coverage. Logic-in-Frames~\cite{guo2025logic} integrates visual–logical dependencies (e.g., causality, spatial relations) to extract semantically rich frames from long videos.

These approaches demonstrate the promise of embedding-aware selection, but most rely on a single criterion—semantic similarity, temporal coverage, or logical reasoning—limiting their ability to generalize across diverse tasks.  

Our \textbf{KeyScore} addresses this by introducing a hybrid scoring scheme that combines three complementary signals:  
(1) \textbf{semantic similarity}, measuring alignment with caption embeddings;  
(2) \textbf{temporal distinctiveness}, encouraging diverse event coverage over time; and  
(3) \textbf{drop impact}, penalizing redundant or low-utility frames.  

This multi-faceted scoring provides a richer assessment of frame importance, yielding more balanced and context-aware selection for downstream retrieval, classification, and summarization tasks.

\section{Method Overview}
Given a raw video, our method aims to efficiently select a small set of semantically informative and temporally diverse keyframes for downstream video-language tasks. The pipeline consists of two main stages: (1) \textbf{STACFP} for frame proposal via spatio-temporal adaptive clustering, and (2) \textbf{KeyScore} for fine-grained frame scoring based on semantic and structural cues.

As illustrated in Figure~\ref{fig:pipeline}, a video is first processed by STACFP to generate candidate frames. These frames are then encoded and evaluated by KeyScore, which integrates semantic similarity, temporal contribution, and drop impact to assign importance scores. A task-dependent thresholding step selects the final keyframes used for retrieval, classification, or summarization.

\subsection{Spatio-Temporal Adaptive Clustering Frame Proposal (STACFP)}
\label{sec:samplers}


Long videos contain thousands of redundant or irrelevant frames, making full-frame processing computationally costly and unnecessary. We propose \textbf{STACFP}, a lightweight unsupervised method that selects a compact set of visually diverse and temporally distributed frames for downstream scoring or inference.

Unlike uniform sampling or prior clustering-based methods like Katna~\cite{katna} and VSUMM~\cite{vsumm}, STACFP encodes both appearance and time in its clustering space. 
For each sampled frame $f_i$, we extract a low-level visual feature vector $\mathbf{v}_i$ based on color histograms computed in HSV color space, which is more perceptually aligned than RGB. This histogram is flattened into a vector of fixed dimension $d$.
To encourage temporal dispersion in the clustering process, we also encode the normalized timestamp of each frame $t_i = \frac{i}{N-1}$, where $i$ is the index of the frame among $N$ total sampled frames. This scalar is then scaled by a hyperparameter $\gamma_{\text{time}}$ and concatenated with the visual feature:
\[
\mathbf{x}_i = [v_i;\, \gamma_{\text{time}} \cdot t_i]
\]
This results in a $(d+1)$-dimensional feature vector $\mathbf{x}_i$ for each frame. The hyperparameter $\gamma_{\text{time}} \in [5, 20]$ controls the influence of temporal position relative to visual appearance in the clustering process.

We perform $k$-means clustering over these spatio-temporal features and automatically select the optimal number of clusters $k^*$ via silhouette score maximization~\cite{silhouettes_score}:
\[
k^* = \arg\max_k \text{Silhouette}(X, \text{KMeans}(k))
\]
This adaptive strategy allocates fewer proposals to static scenes and more to dynamic content. The final frame proposals are chosen as the nearest frames to each cluster centroid. 
This procedure is summarized in Algorithm~\ref{algo:stacfp}.

\vspace{-1mm}
\begin{algorithm}
\caption{STACFP}
\begin{algorithmic}[1]
\State Sample frames every $N$ steps from video
\For{each sampled frame}
    \State Extract low-level visual feature $v_i$ (e.g., HSV histogram)
    \State Normalize timestamp $t_i$
    \State Form spatio-temporal vector $\mathbf{x}_i = [v_i;\, \gamma_{\text{time}} \cdot t_i]$
\EndFor
\State Stack $X = [\mathbf{x}_1, \dots, \mathbf{x}_n]$
\For{$k$ in $[k_{\min}, k_{\max}]$}
    \State Compute KMeans$(X, k)$ and silhouette score
\EndFor
\State Select $k^*$ with highest score
\State Return 1 representative frame per cluster as proposals
\end{algorithmic}
\label{algo:stacfp}
\end{algorithm}
\vspace{-1mm}

\begin{figure}[t]
\centering
\includegraphics[width=\linewidth]{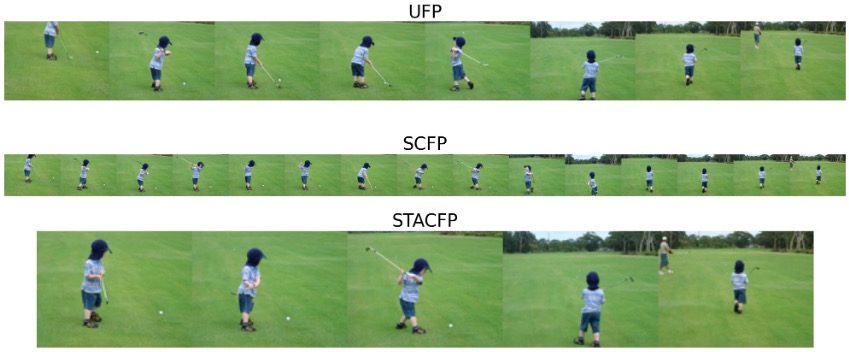}
\caption{\textbf{Qualitative comparison of different frame proposal methods.} 
UFP (Uniform Frame Proposal) samples frames at regular intervals without considering visual or temporal context, often leading to redundancy and suboptimal coverage (e.g., multiple similar frames during the swing motion).
SCFP (Spatial Visual Clustering Frame Proposal) improves diversity via K-means clustering on HSV-based low-level visual features but lacks temporal awareness, resulting in over-sampling static periods.
STACFP (Spatio-Temporal Adaptive Clustering Frame Proposal) combines visual and temporal cues for better coverage of key moments with fewer redundant frames. Notably, STACFP captures the start, middle, and end of the golf swing more effectively, highlighting its ability to preserve action dynamics.
}
\label{fig:sampler_comparison}
\end{figure}

Figure~\ref{fig:sampler_comparison} shows a comparison of different frame proposal methods, including UFP (Uniform Frame Proposal), SCFP (Spatial Clustering Frame Proposal), and STACFP (Spatio-Temporal Adaptive Clustering Frame Proposal). This example illustrates how STACFP better captures key temporal transitions and semantically important moments, while UFP and SCFP often include visually redundant or less informative frames.

\subsection{Frame Scoring via KeyScore}
\label{subsec:KeyScore}



Given a query caption $C$ and a video $V = \{f_1, f_2, \dots, f_T\}$ with $T$ frames, our objective is to estimate the importance of each frame $f_i$ in supporting video–caption alignment. We introduce \textbf{KeyScore}, a hybrid scoring framework that leverages a pretrained video–text model to embed frames and captions into a shared representation space.

Let $\mathbf{f}_i \in \mathbb{R}^D$ denote the embedding of frame $f_i$, $\mathbf{t} \in \mathbb{R}^D$ the embedding of caption $C$, and $\mathbf{v} \in \mathbb{R}^D$ the global video embedding (computed via mean pooling or text-guided attention over $\{\mathbf{f}_i\}$). All embeddings are $\ell_2$-normalized.

\paragraph{Overall scoring.}
KeyScore assigns each frame $f_i$ a weighted score:
\begin{equation}
\text{KeyScore}(i) = \alpha \cdot S_{\text{sem}}(i) + \beta \cdot S_{\text{temp}}(i) + \gamma \cdot S_{\text{drop}}(i)
\end{equation}
where $\alpha + \beta + \gamma = 1$ and each $S_{\cdot}$ captures a complementary aspect of frame importance.

\subsubsection{Semantic Similarity Score ($S_{\text{sem}}$)}
\begin{equation}
S_{\text{sem}}(i) = \cos(\mathbf{f}_i, \mathbf{t})
\end{equation}
$S_{\text{sem}}$ measures how well a frame aligns with the caption.  
\textbf{Example:} For ``a man riding a horse,’’ frames showing the man on horseback obtain higher scores.

\subsubsection{Temporal Representativeness Score ($S_{\text{temp}}$)}
\begin{equation}
S_{\text{temp}}(i) = \cos(\mathbf{f}_i, \mathbf{v})
\end{equation}
$S_{\text{temp}}$ captures how representative a frame is of the overall video context, down-weighting outliers.  
\textbf{Example:} In a cooking tutorial, frames of the chef cooking are representative, while a shot of the wall clock is not.

\subsubsection{Contextual Drop Impact Score ($S_{\text{drop}}$)}
\begin{equation}
S_{\text{drop}}(i) = \cos(\mathbf{v}, \mathbf{t}) - \cos(\widetilde{\mathbf{v}}^{(-i)}, \mathbf{t})
\end{equation}
$S_{\text{drop}}$ measures the \emph{marginal contribution} of frame $f_i$ by measuring how much video–text similarity degrades when the frame is removed. 
A high score indicates that the frame provides indispensable context for aligning the video with the caption, while redundant or uninformative frames yield near-zero impact.  
\textbf{Example:} For ``a woman performs a ballet spin,’’ excluding the spin frame sharply reduces alignment, revealing its critical role.

\paragraph{Implementation.}
All components are min–max normalized before combination. KeyScore can be efficiently computed with vectorized pooling, and returns both raw and weighted scores for downstream selection or ranking.

\begin{figure*}[t]
    \centering
    \begin{subfigure}{0.48\textwidth}
        \centering
        \includegraphics[width=\linewidth]{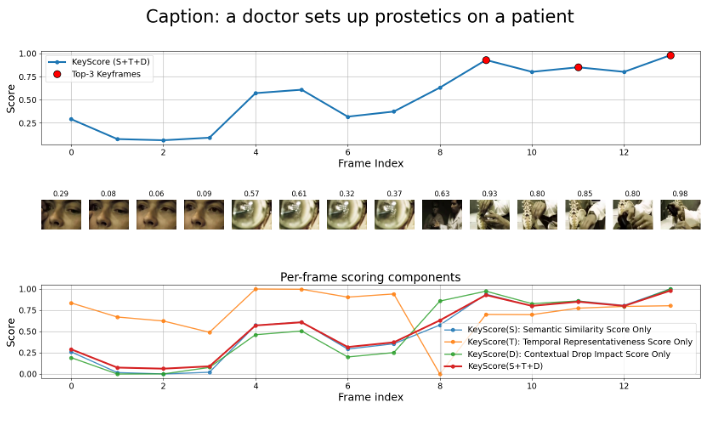}
        \caption{.}
        \label{fig:hyscore_prosthetics}
    \end{subfigure}
    \hfill
    \begin{subfigure}{0.48\textwidth}
        \centering
        \includegraphics[width=\linewidth]{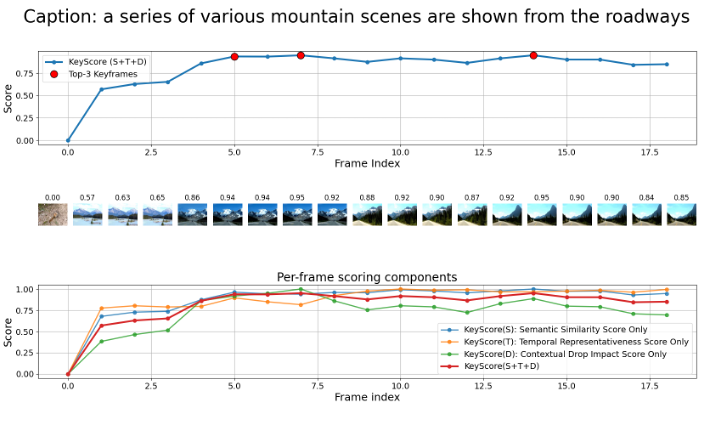}
        \caption{.}
        \label{fig:hyscore_mountains}
    \end{subfigure}
    \vskip 0.2cm
    \begin{subfigure}{0.48\textwidth}
        \centering
        \includegraphics[width=\linewidth]{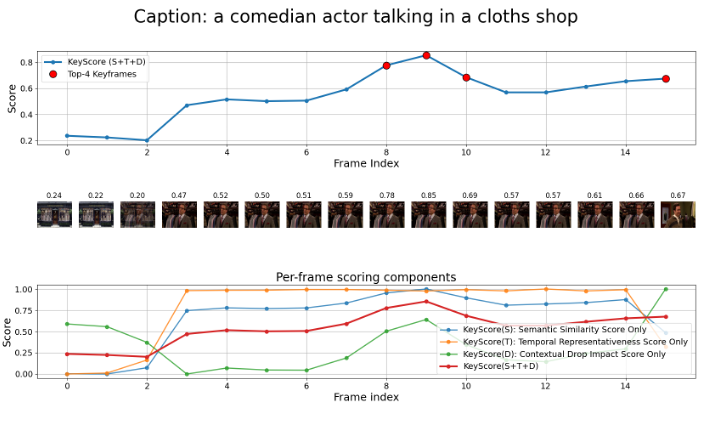}
        \caption{.}
        \label{fig:hyscore_actor}
    \end{subfigure}
    \hfill
    \begin{subfigure}{0.48\textwidth}
        \centering
        \includegraphics[width=\linewidth]{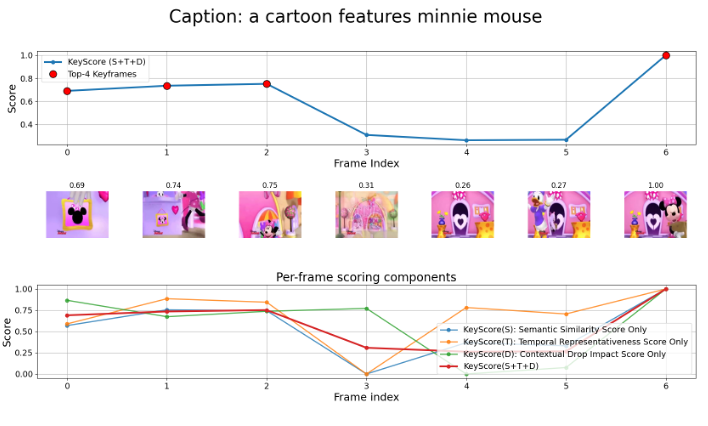}
        \caption{.}
        \label{fig:hyscore_cartoon}
    \end{subfigure}
    \vspace{-3.5mm}
    \caption{\textbf{Qualitative examples of KeyScore frame scoring across diverse scenarios.} 
    Each example shows (top) the final KeyScore curve with top keyframes highlighted, (middle) uniformly sampled frames with scores, and (bottom) individual scoring components. 
    The \textbf{semantic similarity score (S)} reliably highlights frames that directly align with the caption, ensuring semantic grounding (prosthetic setup in (a), mountain landscapes in (b), actor in (c), and Minnie Mouse in (d)). 
    The \textbf{contextual drop impact score (D)} emphasizes indispensable frames whose removal significantly degrades video–text similarity, ensuring key evidence is preserved. 
    The \textbf{temporal representativeness score (T)} favors frequently recurring frames, which supports temporal coverage but may introduce redundancy or less informative content. 
    Combining all three signals yields compact sets of keyframes that maximize semantic relevance and contextual saliency while maintaining temporal diversity. 
    Examples are taken from MSR-VTT~\cite{msrvtt}.}
    \label{fig:KeyScore_examples}
\end{figure*}

Figure~\ref{fig:KeyScore_examples} presents four qualitative examples of KeyScore applied to different video–caption pairs. 
In the prosthetic setup video (Fig.~\ref{fig:hyscore_prosthetics}), KeyScore focuses on frames that visually capture the medical procedure, while down-weighting irrelevant early frames. 
In the mountain scenes video (Fig.~\ref{fig:hyscore_mountains}), most frames align with the caption, and KeyScore identifies representative landscape shots without redundancy. 
The comedian actor example (Fig.~\ref{fig:hyscore_actor}) highlights frames where the actor is clearly visible and contextually important, while the Minnie Mouse cartoon example (Fig.~\ref{fig:hyscore_cartoon}) selects frames where the character appears prominently. 

Across all cases, semantic similarity (S) and contextual drop impact (D) are the strongest contributors, ensuring semantic and contextual fidelity. 
Temporal representativeness (T), although less discriminative, provides complementary coverage by selecting recurring frames. 
Together, these signals enable KeyScore to select just 2–3 frames that faithfully capture the essential visual evidence described by the caption, while discarding redundant or irrelevant content.

\section{Experiments}
We evaluate KeyBench on three downstream tasks: video-text retrieval across three benchmark datasets, zero-shot video action classification, and keyframe extraction on two widely used small-scale benchmarks.

\subsection{Zero-Shot Video-Text Retrieval}
\label{sec:retrieval}
Our evaluation covers four aspects: 
(1) the impact of different frame sampling strategies,  
(2) a comparison of state-of-the-art encoders when combined with our proposed frame scoring,  
(3) the performance of state-of-the-art models against our KeyScore method, and  
(4) the effectiveness of our approach in terms of frame compression rate.

\paragraph{Setup.}
We follow standard evaluation protocols, reporting Recall@K (R@1/5/10) for both text-to-video (T2V) and video-to-text (V2T) retrieval.

\paragraph{Backbone and Input.}
Unless otherwise noted, we use a state-of-the-art video encoder, Perception Encoder (PE)~\cite{PE} as the vision–language backbone. Each video is represented by a small set of keyframes produced by a frame sampler (Sec.~\ref{sec:samplers}). When KeyScore is enabled (Sec.~\ref{sec:KeyScore}), it re-ranks and selects the final subset used for retrieval.

\paragraph{Datasets.}
We report results on MSR-VTT~\cite{msrvtt}, MSVD~\cite{msvd}, DiDeMo~\cite{didemo} following common splits and protocols.

\subsubsection{Frame Samplers}
\label{sec:samplers}
We compare three frame sampling strategies under a controlled retrieval setup:

\begin{itemize}[leftmargin=1.2em]
    \item \textbf{Uniform Frame Proposal (UFP):} Fixed-interval sampling, commonly used in video encoders and Video-LLMs (typically 8 frames). Simple and efficient, but prone to redundancy and sensitive to frame count.
    \item \textbf{SCFP:} KMeans-based clustering in visual space to select diverse keyframes. Reduces redundancy but ignores temporal structure.
    \item \textbf{STACFP (ours):} Spatio-temporal adaptive clustering guided by silhouette analysis. Preserves visual diversity while ensuring temporal coverage, enabling efficient frame selection with fewer samples.  
\end{itemize}
As shown in Table~\ref{tab:retrieval_sampler_msr_ms}, all three methods achieve comparable retrieval accuracy on MSRVTT and MSVD when paired with the same encoder ($\text{PE}_{\text{core}}$G~\cite{PE}). However, STACFP attains this performance with substantially fewer frames—averaging 6 and 5.6 frames per video—compared to 8 for UFP and up to 16 for SCFP. This highlights the efficiency of STACFP, which adaptively selects a compact yet representative subset of frames by integrating temporal cues with clustering quality.  

While SCFP reduces redundancy, its purely visual clustering overlooks temporal coherence. UFP remains competitive but requires careful tuning of the frame budget. In contrast, STACFP consistently offers the best trade-off between retrieval accuracy and frame efficiency, making it a practical choice for scalable video representation.

\begin{table}[t]
\centering
\footnotesize
\setlength{\tabcolsep}{5pt}
\caption{\textbf{Comparison of frame sampling strategies on retrieval performance.} We compare UFP, SCFP~\cite{katna}, and our proposed STACFP, each paired with the same encoder~\cite{PE}. STACFP achieves competitive accuracy with significantly fewer frames than SCFP and UFP, demonstrating its efficiency for video–text retrieval. T2V/V2T: Recall@1 (\%); ASF: average sampled frames.}
\label{tab:retrieval_sampler_msr_ms}
\begin{tabular}{lccc|ccc}
\toprule
\multirow{2}{*}{Frame Sampler} & \multicolumn{3}{c|}{\textbf{MSRVTT}} & \multicolumn{3}{c}{\textbf{MSVD}} \\
\cmidrule(lr){2-4} \cmidrule(lr){5-7}
 & T2V & V2T & ASF & T2V & V2T & ASF \\
\midrule
UFP & 50.0 & \textbf{71.7} & 8.0 & 60.4 & \textbf{82.9} & 8.0 \\
SCFP & 49.4 & 71.3 & 16.0 & 59.9 & 82.3 & 10.7 \\
\textbf{STACFP} & \textbf{49.7} & 71.1 & \textbf{6.0} & \textbf{60.4} & 82.3 & \textbf{5.6} \\
\bottomrule
\end{tabular}
\end{table}

\subsubsection{KeyScore: Frame Scoring and Selection}
\label{sec:KeyScore}

Given initial frame proposals from STACFP, we further score each frame using \textbf{KeyScore}, a weighted combination of three complementary cues: semantic similarity (S), temporal representativeness (T), and contextual drop impact (D). 

Table~\ref{tab:KeyScore_ablation} presents an ablation across MSRVTT~\cite{msrvtt}, MSVD~\cite{msvd}, and DiDeMo~\cite{didemo}, comparing individual and joint scoring signals. The PE-only baseline uses 6–11 frames per video and yields modest retrieval performance. Adding KeyScore significantly improves retrieval accuracy while substantially reducing the number of frames.

Among single signals, semantic similarity (S) and contextual drop impact (D) are the most effective, boosting MSRVTT T2V R@1 above 62 and DiDeMo above 77. Temporal representativeness (T) alone contributes little, but enhances performance when combined with other signals. Pairwise combinations like KeyScore(S+D) already deliver strong gains across datasets. 

The best results are obtained with the full combination KeyScore(S+T+D), achieving 63.9/60.5 R@1 on MSRVTT, 89.2/89.2 on MSVD, and 80.7/89.2 on DiDeMo — all while using only 2–2.5 frames on average. This demonstrates KeyScore’s ability to balance semantic, temporal, and contextual factors for compact yet informative frame selection.

\begin{table*}[t]
\centering
\footnotesize
\caption{Ablation of KeyScore frame scoring on MSRVTT, MSVD, and DiDeMo.
We report text-to-video (T2V) and video-to-text (V2T) R@1 retrieval accuracy (\%) and average selected frames (ASF).
S: semantic similarity, $S_{\text{sem}}$, T: temporal representativeness, $S_{\text{temp}}$, D: contextual drop impact, $S_{\text{drop}}$.}
\label{tab:KeyScore_ablation}
\begin{tabular}{l|ccc|ccc|ccc}
\toprule
\multirow{2}{*}{Method} & \multicolumn{3}{c|}{MSRVTT} & \multicolumn{3}{c|}{MSVD} & \multicolumn{3}{c}{DiDeMo} \\
& T2V & V2T & ASF & T2V & V2T & ASF & T2V & V2T & ASF \\
\midrule
PE\textsubscript{core}G (STACFP only) & 49.7 & 47.8 & 6   & 60.4 & 82.9 & 5.6 & 44.0 & 46.1 & 11 \\
PE\textsubscript{core}G + KeyScore(S)         & 63.2 & 60.0 & 2   & 88.5 & 86.5 & 5   & 79.3 & 86.5 & 3 \\
PE\textsubscript{core}G  + KeyScore(T)         & 49.8 & 47.9 & 8 & 84.6 & 86.1 & 4   & 71.5 & 86.1 & 2 \\
PE\textsubscript{core}G  + KeyScore(D)         & 62.6 & 59.4 & 3   & 85.8 & 86.5 & 3   & 77.6 & 86.5 & 2 \\
PE\textsubscript{core}G  + KeyScore(S+T)       & 61.3 & 59.5 & 3 & 87.9 & 88.6 & 2   & 80.8 & 88.6 & 2 \\
PE\textsubscript{core}G  + KeyScore(D+T)       & 61.4 & 59.1 & 2 & 87.9 & 89.2 & 4   & 80.6 & 89.2 & 2 \\
PE\textsubscript{core}G  + KeyScore(S+D)       & 63.5 & 60.3 & 2 & 89.1 & 89.7 & 2   & 80.5 & 89.7 & 2 \\
\textbf{PE\textsubscript{core}G  + KeyScore(S+T+D)} & \textbf{63.9} & \textbf{60.5} & \textbf{2.5} & \textbf{89.2} & \textbf{89.2} & \textbf{2} & \textbf{80.7} & \textbf{89.2} & \textbf{2} \\
\bottomrule
\end{tabular}
\end{table*}
vspace{-2mm}

\subsubsection{Comparison with State of the Art}
As shown in Section~\ref{sec:KeyScore}, integrating our KeyScore into the retrieval pipeline substantially boosts performance by filtering out redundant frames and preserving only the most informative ones. This improves the overall quality of frame inputs before they are processed by the encoder.

We compare PE with and without KeyScore against recent state-of-the-art video–language models. Retrieval performance is reported using Recall@1 (R@1) for both text-to-video (T2V) and video-to-text (V2T), as summarized in Table~\ref{tab:retrieval_results}.

The results demonstrate that PE\textsubscript{core}G with KeyScore achieves new state-of-the-art results across all three benchmarks. On MSRVTT, it improves R@1 to 63.9\% (T2V) and 60.5\% (V2T), surpassing InternVideo2 and other strong baselines. On MSVD, the gains are even larger, reaching 89.2\% for both tasks, significantly outperforming models such as SigLIP2 and VideoPrism-g. On DiDeMo, a particularly challenging dataset, our method attains 60.4\% (T2V) and 60.3\% (V2T), again outperforming prior approaches. These results highlight the effectiveness of KeyScore in enhancing cross-modal alignment and retrieval robustness across diverse benchmarks.

\begin{table}[t]
\centering
\caption{\textbf{Zero-shot video–text retrieval (R@1)} on MSRVTT, MSVD, and DiDeMo.  
Results are reported for text-to-video (T2V) and video-to-text (V2T).  
PE\textsubscript{core}G augmented with KeyScore achieves consistent state-of-the-art gains across all datasets.}
\resizebox{\columnwidth}{!}{%
\begin{tabular}{lcccccc}
\toprule
\multirow{2}{*}{Model} & \multicolumn{2}{c}{MSRVTT} & \multicolumn{2}{c}{MSVD} & \multicolumn{2}{c}{DiDeMo} \\
\cmidrule(lr){2-3} \cmidrule(lr){4-5} \cmidrule(lr){6-7}
 & T2V & V2T & T2V & V2T & T2V & V2T \\
\midrule
CLIP~\cite{clip} & 30.4 & 24.2 & 40.5 & 57.2 & 12.7 & -- \\
CLIP4Clip~\cite{clip4clip} & 32.0 & --   & 45.2 & 48.4 & --   & -- \\
X-CLIP~\cite{xclip} & 49.3 & 48.9 & 50.4 & 66.8 & 47.8 & 47.8 \\
UMT-L~\cite{umt-l} & 40.7 & 37.1 & 49.0 & 74.5 & 49.9 & 59.7 \\
SigLIP2-L/16~\cite{siglip} & 41.5 & 31.4 & 53.7 & 74.2 & 18.4 & -- \\
InternVL~\cite{internvl} & 44.7 & 40.2 & 43.4 & 67.6 & --   & -- \\
ViCLIP~\cite{viclip} & 42.4 & 41.3 & 49.1 & 75.1 & 31.5 & -- \\
InternVideo2~\cite{internvideo2} & 51.9 & 50.9 & --   & --   & 57.9 & 57.1 \\
VideoPrism-g~\cite{videoprism} & 39.7 & 71.0 & 58.1 & 83.3 & --   & -- \\
SigLIP2-g-opt~\cite{siglip} & 43.1 & 34.2 & 55.8 & 74.6 & --   & -- \\
\midrule
PE\textsubscript{core}G (image)~\cite{PE} & 44.3 & 35.2 & 54.3 & 73.9 & --   & -- \\
PE\textsubscript{core} (video) & 51.2 & 49.9 & 59.7 & 85.4 & 43.1 & 45.1 \\
\textbf{PE\textsubscript{core}G (KeyScore)} & \textbf{63.9} & \textbf{60.5} & \textbf{89.2} & \textbf{89.2} & \textbf{60.4} & \textbf{60.3} \\
\bottomrule
\end{tabular}%
}
\label{tab:retrieval_results}
\end{table}

\subsubsection{Frame Reduction Analysis}
To evaluate the efficiency of KeyScore, we measure how many frames are discarded relative to common baselines. 
We introduce the \emph{Frame Reduction Rate} (FRR), defined as the proportion of frames saved:

\[
\text{FRR-UFP} = 1 - \frac{N_{\text{sel}}}{N_{\text{UFP}}}, \qquad
\text{FRR-Avg} = 1 - \frac{N_{\text{sel}}}{N_{\text{avg}}},
\]
where \(N_{\text{sel}}\) is the number of frames selected by KeyScore, 
\(N_{\text{UFP}}=8\) is the standard uniform fixed sampling baseline, 
and \(N_{\text{avg}}\) is the dataset-specific average frame count. 
Higher FRR indicates greater reduction, i.e., more frames are saved.

Table~\ref{tab:KeyScore_frr_analysis} reports ASF, FRR-UFP, and FRR-Avg across datasets for different KeyScore variants.

\paragraph{MSRVTT.}  
On MSRVTT (avg. 408 frames), KeyScore reduces inputs to 2–3 frames, giving \textbf{FRR-Avg $\approx$ 0.99} and \textbf{FRR-UFP = 0.59–0.75}. The full S+D+T variant selects 2.5 frames with FRR-UFP = 0.69 and FRR-Avg = 0.99.

\paragraph{MSVD.}  
On MSVD (avg. 275 frames), KeyScore consistently selects 2 frames in its strongest settings (\textbf{FRR-UFP = 0.75}, \textbf{FRR-Avg = 0.99}), and never exceeds 6 frames across all variants.

\paragraph{DiDeMo.}  
On DiDeMo (avg. 275 frames), KeyScore reduces 11 frames to 2–3 (\textbf{FRR-Avg $\approx$ 0.99}, \textbf{FRR-UFP = 0.63–0.75}), showing robustness on longer, diverse videos.

\paragraph{Conclusion.}  
KeyScore consistently saves over \textbf{70\% vs. UFP} and nearly \textbf{99\% vs. dataset averages}, retaining only 2–3 informative frames per video. The S+D+T variant offers the best balance, confirming the complementarity of semantic, temporal, and drop-aware scoring.
\begin{table*}[t]
\centering
\footnotesize
\setlength{\tabcolsep}{6pt}
\caption{\textbf{Analysis of KeyScore combinations across datasets.} We report Average Selected Frames (ASF), FRR-UFP (drop-resilience), and FRR-Avg (overall frame retention reliability). Combining semantic (S), temporal (T), and drop impact (D) scores yields the best trade-off between robustness and efficiency.}
\begin{tabular}{lccc|ccc|ccc}
\toprule
\multirow{2}{*}{Frame Scoring} &
\multicolumn{3}{c}{\textbf{MSRVTT (avg. 408 frames)}} &
\multicolumn{3}{c}{\textbf{MSVD (avg. 275 frames)}} &
\multicolumn{3}{c}{\textbf{DiDeMo (avg. 1728 frames)}} \\
\cmidrule(lr){2-4} \cmidrule(lr){5-7} \cmidrule(lr){8-10}
& ASF & FRR-UFP$^{\uparrow}$ & FRR-Avg$^{\uparrow}$ 
& ASF & FRR-UFP$^{\uparrow}$ & FRR-Avg$^{\uparrow}$ 
& ASF & FRR-UFP$^{\uparrow}$  & FRR-Avg$^{\uparrow}$  \\
\midrule
PE\textsubscript{core}G + KeyScore(S)      & 2.00 & 0.75   & 0.99   & 5.00 & 0.38 & 0.98   & 3.00 & 0.63 & 0.99 \\
PE\textsubscript{core}G + KeyScore(T)      & 8.20 & -0.03  & 0.98   & 4.00 & 0.50 & 0.99   & 2.00 & 0.75 & 0.99 \\
PE\textsubscript{core}G + KeyScore(D)      & 3.00 & 0.63   & 0.99   & 6.00 & 0.25 & 0.98   & 2.00 & 0.75 & 0.99 \\
PE\textsubscript{core}G + KeyScore(S+T)    & 3.30 & 0.59   & 1.00   & 2.00 & 0.75 & 0.99   & 2.00 & 0.75 & 0.99 \\
PE\textsubscript{core}G + KeyScore(D+T)    & 2.57 & 0.68   & 0.99   & 2.00 & 0.50 & 0.99   & 2.00 & 0.75 & 0.99 \\
PE\textsubscript{core}G + KeyScore(S+D)    & 2.69 & 0.66   & 0.99   & 2.00 & 0.75 & 0.99   & 2.00 & 0.75 & 0.99 \\
\textbf{PE\textsubscript{core}G + KeyScore(S+D+T)} & \textbf{2.50} & \textbf{0.69} & \textbf{0.99} & \textbf{2.00} & \textbf{0.75} & \textbf{0.99} & \textbf{2.00} & \textbf{0.75} & \textbf{0.99} \\
\bottomrule
\end{tabular}
\label{tab:KeyScore_frr_analysis}
\end{table*}


\subsection{Keyframe Extraction}

We evaluate KeyScore on two widely used keyframe extraction benchmarks: TVSum20~\cite{tvsum} and SumMe~\cite{summe}. 
For TVSum, we pair KeyScore with CLIP-ViT-H/14~\cite{clip}, while for SumMe we use PE\textsubscript{core}G~\cite{PE} with KeyScore. 
Following the evaluation protocol of~\cite{cakmak2025tripss}, we report F1 scores computed with frame-level color histogram similarity. 
As shown in Table~\ref{tab:tvsum_summe}, KeyScore and its variants achieve strong results, outperforming TRIPSS\textsubscript{semantic} and several recent baselines despite relying solely on semantic alignment. 
In contrast to TRIPSS~\cite{cakmak2025tripss}, which integrates semantic, perceptual, and structural cues, \name focuses on semantic relevance tailored to downstream tasks, demonstrating that semantic-guided scoring alone can effectively identify keyframes for both model consumption and human interpretation.

\begin{table}[t]
\footnotesize
\centering
\caption{F1 scores on \textbf{TVSum20}~\cite{tvsum} and \textbf{SumMe}~\cite{summe}. 
CLIP/PE combined with our KeyScore consistently outperforms or matches existing baselines.}
\vspace{-0.3cm}
\begin{tabular}{l c | l c}
\toprule
\multicolumn{2}{c}{\textbf{TVSum20}} & \multicolumn{2}{c}{\textbf{SumMe}} \\
\midrule
\textbf{Method} & \textbf{F1-Score} & \textbf{Method} & \textbf{F1-Score} \\
\midrule
Uniform & 0.165 & Uniform & 0.283 \\
HistDiff~\cite{histdiff} & 0.338 & H-MAN~\cite{hman} & 0.518 \\
VS-UID~\cite{garcia2023videosum} & 0.462 & SUM-GDA~\cite{sumgda} & 0.528 \\
GMC~\cite{gmc} & 0.483 & STVS~\cite{kashid2024stvs} & 0.536 \\
VSUMM~\cite{vsumm} & 0.489 & TAC-SUM~\cite{tacsum} & 0.545 \\
KMKey~\cite{kmkey} & 0.504 & PGL-SUM~\cite{pglsum} & 0.556 \\
LBP-Shot~\cite{nandini2022shot} & 0.505 & SMN~\cite{smn} & 0.583 \\
VS-Inception~\cite{garcia2023videosum} & 0.517 & AugFusion~\cite{augfusion} & 0.584 \\
LMSKE~\cite{LMSKE} & 0.531 & Ldpp-c~\cite{ldppc} & 0.588 \\
TRIPSS\textsubscript{semantic}~\cite{cakmak2025tripss}  & 0.487 & TRIPSS\textsubscript{semantic} & 0.430 \\
TRIPSS & 0.610 & TRIPSS & 0.590 \\
\midrule
\textbf{CLIP~\cite{clip} + KeyScore} & \textbf{0.539} & \textbf{PE~\cite{PE} + KeyScore} & \textbf{0.655} \\
\bottomrule
\end{tabular}
\label{tab:tvsum_summe}
\end{table}

\subsection{Zero-Shot Video Action Classification}
We further evaluate our frame proposal and scoring strategies on the HMDB-51~\cite{hmdb51} benchmark, which contains 51 human action categories. 
Following Qwen-2.5-VL~\cite{qwen2}, we first generate captions for each video clip and use them to guide KeyScore-based frame scoring. 
For classification, we employ the PE\textsubscript{core}G~\cite{PE} frame-based video encoder. 
Frames are selected according to score thresholds, and for scoring-based methods we report the best F1 obtained across thresholds.  

Table~\ref{tab:hmdb51} presents zero-shot video action classification results on HMDB51. 
Among the baseline models, InternVL~\cite{internvideo}, InternVideo2~\cite{internvideo2}, and SigLIP2-g-opt~\cite{tschannen2025siglip} achieve F1 scores in the 0.518–0.555 range with FRR-Avg values of 0.915, reflecting strong but comparable performance across different architectures and resolutions.  

In contrast, PE\textsubscript{core}G + KeyScore delivers a substantial improvement, achieving an F1 of \textbf{0.675} and an FRR-Avg of \textbf{0.972}. 
This represents an absolute gain of +12.0 F1 points over the strongest baseline (InternVL), while simultaneously discarding a larger fraction of frames. 
The higher FRR-Avg demonstrates that KeyScore can aggressively reduce frame inputs while preserving the frames most critical for action understanding.  

These results reveal two important trends. 
First, semantic- and context-aware scoring is more effective for action classification than dense uniform sampling, as KeyScore prioritizes frames aligned with action semantics rather than treating all frames equally. 
Second, KeyScore’s ability to retain fewer frames yet improve accuracy highlights its efficiency, making it particularly suitable for large-scale video understanding tasks where both performance and computational cost are critical.  
Overall, the combination of PE\textsubscript{core}G with KeyScore establishes a new state of the art on HMDB51 under zero-shot evaluation by jointly optimizing recognition accuracy and frame efficiency.


\begin{table}[t]
\footnotesize
\centering
\setlength{\tabcolsep}{6pt}
\renewcommand{\arraystretch}{1.2}
\caption{Zero-shot video action classification results on \textbf{HMDB51}~\cite{hmdb51}. Our method (PE\textsubscript{core}G + KeyScore) achieves the best F1 with the highest FRR-Avg.}
\vspace{-0.1in}
\begin{tabular}{l|c c c}
\toprule
\textbf{Model} & \textbf{Resolution} & \textbf{F1-Score} & \textbf{FRR-Avg$^{\uparrow}$} \\
\midrule
InternVL~\cite{internvideo}            & 224 & 0.555 & 0.915 \\
InternVideo2~\cite{internvideo2}       & 224 & 0.539 & 0.915 \\
SigLIP2-g-opt~\cite{tschannen2025siglip} & 384 & 0.518 & 0.915 \\
\textbf{PE\textsubscript{core}G~\cite{PE} + KeyScore} & \textbf{448} & \textbf{0.675} & \textbf{0.972} \\
\bottomrule
\end{tabular}
\label{tab:hmdb51}
\end{table}

\section{Conclusion}
\label{sec:conclusion}
We introduced \textbf{KeyScore}, a caption-aware frame scoring method that integrates semantic similarity, temporal representativeness, and contextual drop impact to identify the most informative frames in a video. 
Unlike traditional uniform sampling or clustering-based approaches, KeyScore leverages video–caption alignment to filter out uninformative frames while preserving semantic diversity.  


Experiments on retrieval, summarization, and action recognition benchmarks show that KeyScore improves accuracy while cutting frame usage by 70–99\% over full videos and 63–75\% over 8-frame baselines.

By generating frame-level importance scores from video–caption pairs, KeyScore offers a scalable tool for keyframe extraction, video encoder, and Video-LLMs, laying the foundation for content-efficient video understanding and semantics-driven frame selection

{
    \small
    \bibliographystyle{ieeenat_fullname}
    \bibliography{main}
}

\end{document}